\documentclass[runningheads]{llncs}
\usepackage[T1]{fontenc}
\usepackage{graphicx}
\usepackage{booktabs}
\usepackage[misc]{ifsym}
\newcommand{\corr}{(\Letter)}
\usepackage{mwe}
\usepackage{amssymb}
\usepackage{amsmath}
\usepackage{enumitem}
\usepackage{color}
\usepackage{times}
\usepackage{soul}
\usepackage{url}
\usepackage[utf8]{inputenc}
\usepackage{amsfonts}
\usepackage{algorithm}
\usepackage{algorithmic}
\usepackage[normalem]{ulem}
\usepackage{multirow}
\usepackage{tabularray}
\usepackage{ulem}
\usepackage{hyperref}

\begin{document}

\title{BotTrans: A Multi-Source Graph Domain Adaptation Approach for Social Bot Detection}

\titlerunning{BotTrans}

\author{Author information scrubbed for double-blind reviewing}
\author{Boshen Shi\inst{1,2} \and
Yongqing Wang\inst{1} \corr \and
Fangda Guo\inst{1}\and Jiangli Shao\inst{1,2} \and Huawei Shen\inst{1,2} \and Xueqi Cheng\inst{1,2}}

\authorrunning{B. Shi et al.}

\institute{ State Key Laboratory of AI Safety, Institute of Computing Technology, Chinese Academy of Sciences, Beijing 100190, China \email{wangyongqing@ict.ac.cn}
\and
University of Chinese Academy of Sciences, Beijing 100190, China \email{shiboshen15@mails.ucas.ac.cn}
}

\toctitle{BotTrans: A Multi-Source Graph Domain Adaptation Approach for Social Bot Detection}
\tocauthor{Boshen Shi, Yongqing Wang, Fangda Guo, Jiangli Shao, Huawei Shen, Xueqi Cheng}

\maketitle              

\begin{abstract}
Transferring extensive knowledge from relevant social networks has emerged as a promising solution to overcome label scarcity in detecting social bots and other anomalies with GNN-based models. However, effective transfer faces two critical challenges. Firstly, the network heterophily problem, which is caused by bots hiding malicious behaviors via indiscriminately interacting with human users, hinders the model's ability to learn sufficient and accurate bot-related knowledge from source domains. Secondly, single-source transfer might lead to inferior and unstable results, as the source network may embody weak relevance to the task and provide limited knowledge. To address these challenges, we explore multiple source domains and propose a multi-source graph domain adaptation model named \textit{BotTrans}. We initially leverage the labeling knowledge shared across multiple source networks to establish a cross-source-domain topology with increased network homophily. We then aggregate cross-domain neighbor information to enhance the discriminability of source node embeddings. Subsequently, we integrate the relevance between each source-target pair with model optimization, which facilitates knowledge transfer from source networks that are more relevant to the detection task. 
Additionally, we propose a refinement strategy to improve detection performance by utilizing semantic knowledge within the target domain. Extensive experiments on real-world datasets demonstrate that \textit{BotTrans} outperforms the existing state-of-the-art methods, revealing its efficacy in leveraging multi-source knowledge when the target detection task is unlabeled. 

\keywords{Graph domain adaptation  \and Social bot detection \and Transfer learning.}
\end{abstract}

\section{Introduction}

Social bots are typically defined as automated social media profiles created to perform specific functions, such as data collection, content sharing, promotion, and activities aimed at political influence~\cite{botsurvey,botsurvey2}. As artificial intelligence advances, social bots increasingly demonstrate the ability to mimic complex aspects of human social behavior. 
They can manipulate public discussions by spreading misinformation, fabricating artificial environments, and even influencing the thoughts and convictions of users~\cite{botgnn_gc1}. As a result, the significance of identifying social bots is progressively growing.

Many attempts at bot detection extract discriminative patterns from user metadata, tweet content and action sequences with shallow and deep methods. Recently, models based on Graph Neural Networks (GNNs) have demonstrated several advantages in social bot detection, especially in capturing and utilizing complex relationships and structures~\cite{botgnn1,botgnn2}. However, due to the evolution of bot strategies and the complex behaviors of bots, they face challenges stemming from the scarcity of annotations~\cite{botsurvey}. Therefore, to detect bots or other anomalies in the challenging unlabeled scenarios, researchers have proposed GNN-based knowledge transfer methods, in which graph domain adaptation (GDA) appears as an effective paradigm~\cite{act,cmd,gdn,grade,adagcn}. The underlying premise is that \textbf{social bots exhibit consistent features and behavioral patterns across social platforms}. With the domain adaptation technique, GDA models can effectively learn bot-related knowledge from task-relevant graphs with abundant bot labels (source domains) and transfer it to task graphs (target domains).

\begin{figure}[h]
  \centering
  \includegraphics[width=0.7\linewidth]{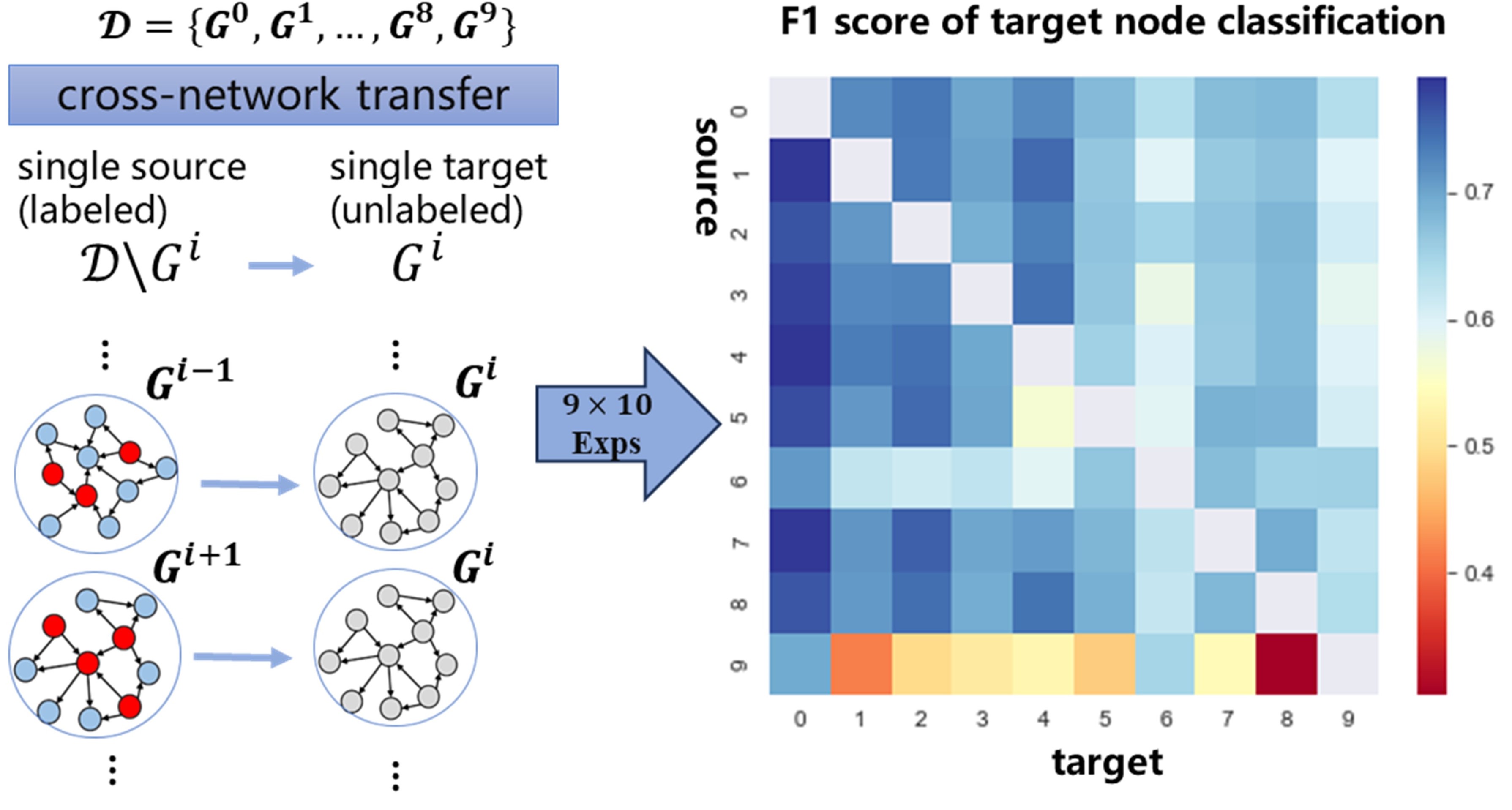}
  \caption{Network-relevance by conducting 9$\times$10 cross-network transfer tasks between every two networks.}
  \label{figidea}
\end{figure}

Although promising, transferring knowledge for detecting social bots faces two challenges which previous studies have largely ignored. 
Firstly, \textbf{the social bot-related knowledge learned from source domains could be insufficient and inaccurate}. As bots frequently interact with human users to conceal their malicious activities, network heterophily arises, i.e. nodes with dissimilar features and labels are connected~\cite{botgnn5,botgnn7}. Aggregating information in a strongly heterophilous neighborhood makes GNNs learn less discriminative node embeddings, especially for social bots. Therefore, ignoring this challenge will raise source training errors and increase classification risk on target domains. 

Another key challenge is that \textbf{transferring knowledge from a single source domain may lead to inferior and unstable adaptation performance}. Prior works have mainly considered transferring from a single source domain,
which may exhibit low relevance to the task and provide limited bot-related knowledge for the model.To facilitate a better understanding, we conducted 9$\times$10 single-source\&single-target transfer experiments with a typical GDA model~\cite{adagcn} over ten domains, each of which represents a different social network with social bots and human users (see Section~\ref{exp-setting} for details). We quantify network relevance based on classification performance. If transferring from source $i$ to target $j$ yields higher accuracy than from source $k$, we infer that $i$ is more relevant to $j$ than $k$. Figure~\ref{figidea} reveals that when the target domain is fixed, the transfer results from different source domains are inconsistent. Similarly, the relevance of the same source domain to different target domains also varies, which is hard to predict prior to implementation. 

To address the above challenges, we explore multiple source graph domains and propose a novel multi-source graph domain adaptation (MSGDA) model, \textit{BotTrans}, to address the unlabeled detection task. Generally, \textit{BotTrans} develops three key modules to guarantee the effectiveness of multi-source knowledge transfer:
\begin{enumerate}
    \item \textbf{Cross Source Domain Message-Passing (CSD-MP).} It alleviates network heterophily problem and enables the model to learn social-bot knowledge more sufficiently from sources. The key idea is sharing labeling knowledge across multiple sources to create a neighborhood with higher homophily for each source node. Specifically, a cross-source-domain topology is established to connect each source node to similar nodes that may be beneficial (e.g., nodes with the same labels) and originate from other source domains. Such cross-source-domain topology embodies higher homophily, and we utilize the aggregated node information from it to improve the discriminative power of source embeddings.
    \item \textbf{Selective Multi-Source Transfer (SMST).} It integrates abundant knowledge from multiple source domains and selectively transfers the most task-relevant knowledge. In particular, we leverage domain-level similarity between each source domain and target domain as an indicator of the relevance of that source domain to the task. Such similarity is computed from both semantic and structural domain information. Subsequently, we apply a weighting strategy during optimization to guide the model in transferring more knowledge from source domains that exhibit higher task relevance to the task.
    \item \textbf{Anomaly Refinement (AR).} It further utilizes the semantic knowledge of the target domain to improve detection performance. A potential risk of knowledge transfer is that models trained on source domains may under-fit on the target domain, which is caused by noisy source labels, i.e., bots are mislabeled as human users. Considering the difficulties in identifying noisy source labels and the missing of target labels, we resort to refining inferior target predictions with the node-level heterophily score. 
\end{enumerate}

Overall, these modules follow the stages of knowledge transfer, which include learning on sources, transferring knowledge to the target, and exploring additional target knowledge. To sum up, we summarize the main contributions of this work as follows:
\begin{enumerate}
\item To the best of our knowledge, we are among the first to study the problem of applying multi-source graph domain adaptation to social bot detection, with the aim of alleviating the scarcity of social bot labels.
\item We propose a novel MSGDA model \textit{BotTrans}. It significantly improves detection performance by alleviating network heterophily on source domains, transferring extensive and task-relevant knowledge to target domains, and refining target detection results.
\item We conduct extensive experiments based on real-world social bot detection datasets, and the results demonstrate that \textit{BotTrans} achieves state-of-the-art performance in unlabeled detection tasks.
\end{enumerate}

\section{Related Works}
\textbf{GNN-Based Social Bot Detection.}
Recently, researchers have resorted to GNN-based models to detect social bots~\cite{botgnn1,botgnn2,botgnn3,botgnn5,botgnn_gc1,botgnn_gc2}. Firstly, as social relationships are more complicated to camouflage, exploring complex social relationships could reveal more discriminative patterns of bots~\cite{relationship1,botrgcn}. Secondly, bots tend to act in groups to pursue malicious goals and implicitly form bot communities~\cite{botsurvey}. Many studies focus solely on addressing label scarcity or network heterophily problems within the task graph, ignoring the potential benefits of exploring auxiliary knowledge from related social platforms with richer annotations, which indicates the importance of developing \textit{BotTrans}. 

\noindent\textbf{Graph Domain Adaptation.}
To overcome the label scarcity problem on graphs, researchers have developed graph domain adaptation (GDA) as an effective transfer learning paradigm on graphs~\cite{gdasurvey}. It typically involves integrating GNNs with domain adaptation techniques to learn network-invariant and label-related node embeddings for graph-related tasks~\cite{shen2023domain,jhgda,specgda,grade,adagcn,asn,udagcn,shi2023node}. GDA has been extended to cross-domain anomaly detection tasks by designing additional topology reconstruction module (Commander~\cite{cmd}) or anomaly-specific objectives for learning social bot-related knowledge on source domains (ACT~\cite{act}). However, they may not be optimal solutions for detecting social bots, as they consider neither the network heterophily problem nor knowledge from multiple sources. Currently, very few works have attempted to resolve the challenging multi-source graph domain adaptation tasks. Representative works include NESTL~\cite{nestl}, MSDS~\cite{msds}, and GDN~\cite{gdn}. However, they either oversimplifies source-target relevance or relies on target labels. Consequently, we develop \textit{BotTrans} to effectively transfer knowledge from multiple source graphs for detecting social bots in the challenging unlabeled scenario.

\section{Problem Formulation}
In this work, we formulate the unlabeled social bot detection task as a \textbf{cross-graph node classification problem}. Specifically, given source graphs $(G^{S_1},\cdots,G^{S_m})$ with abundant labels as source domains, the objective is to learn a detection model transferring labeling knowledge from multiple source domains to predict whether the nodes in the target graph $G^T$ is a bot or not. Each source graph includes a node feature matrix, an adjacency matrix, and node-level labels indicating whether a node represents a bot or not. In contrast, $G^T$ is entirely unlabeled. All graphs share the same feature space.

\section{Methodology}
\begin{figure*}[ht]
  \centering
  \includegraphics[width=0.8\textwidth]{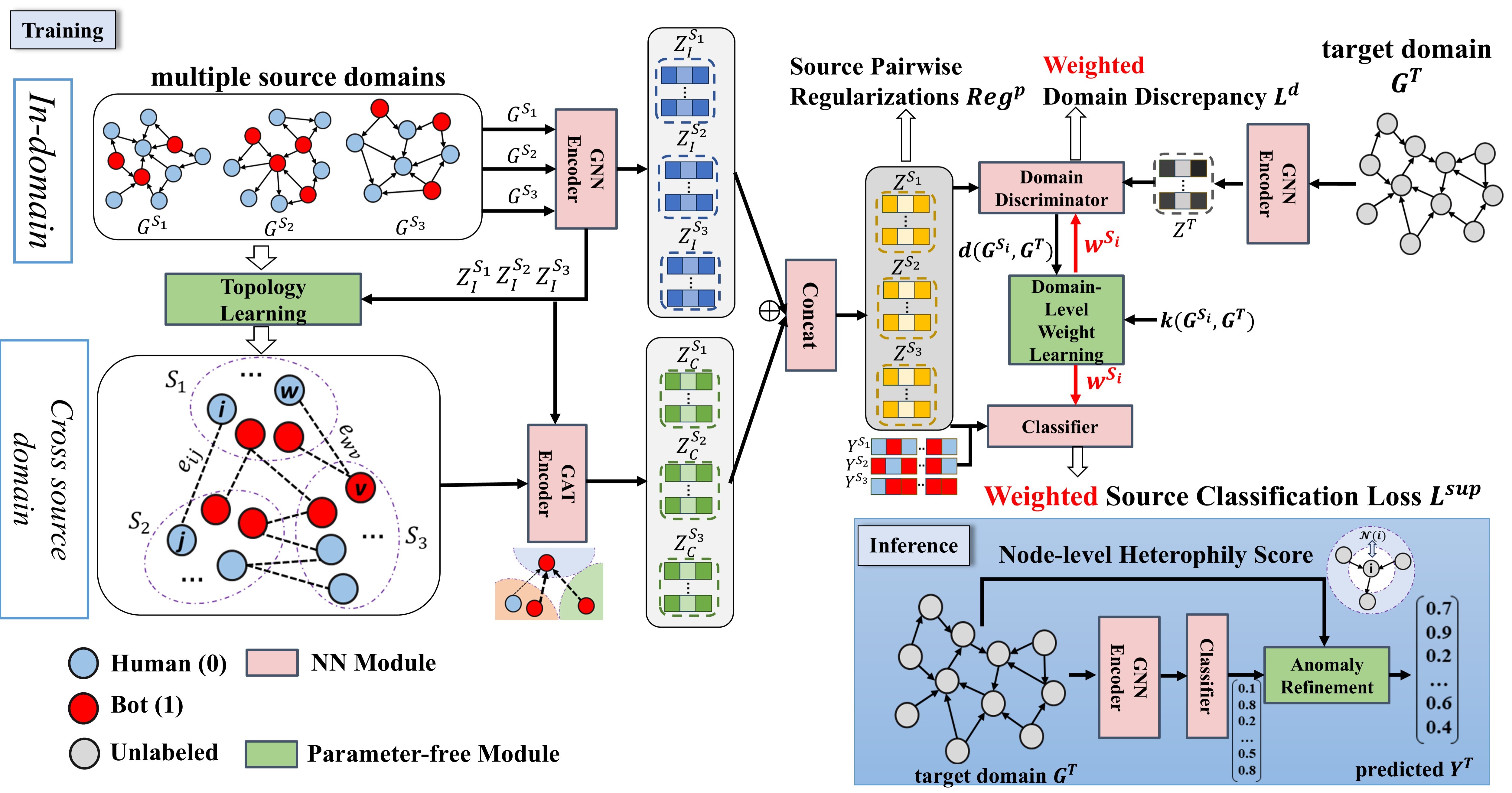}
  \caption{Overview of \textit{BotTrans} framework.}
  \label{figmodel}
\end{figure*}

The framework of \textit{BotTrans} is illustrated in Fig.\ref{figmodel}. Initially, the CSD-MP module generates discriminative source node embeddings by integrating in-domain and cross-source-domain information. Subsequently, the SMST module trains a shared classifier across multiple source domains and adapts it to the target domain with a domain-level weighting strategy. Finally, the AR module refines the classification results output by the under-fitted source classifier during the inference stage. We discuss the details of \textit{BotTrans} in the following sections.

\subsection{Cross Source Domain Message-Passing}
Generally, the CSD-MP module generates node embeddings for all source and target domains. It additionally enhances the discriminability of source node embeddings to facilitate the transfer of more bot-related knowledge. Besides, it aligns the latent embedding spaces of multiple source domains, promoting more stable adaptation from multiple sources to the target.

Given $m$ source graphs $(G^{S_1},\cdots,G^{S_m})$ and a target graph $G^T$, we first utilize a shared GNN encoder $f$ to generate node embeddings $Z_I$ via message-passing inside each network, which is called in-domain message-passing. Take the $l^{th}$ layer of GCN~\cite{gcn} as an example, the in-domain message-passing for computing the embedding of node $i$ from domain $D$ is:
\begin{equation}
\label{indomainMP}
    Z^{D,(l)}_{I,i} = \sigma(b^{(l-1)} + \sum_{j\in\mathcal{N}(i)\cup i}\frac{1}{c_{ij}}Z_{I,j}^{D,(l-1)}W^{(l-1)}) 
\end{equation}
where $D\in \{S_1,\cdots, S_m, T\}$, $b^{(l-1)}$ and $W^{(l-1)}$ are parameters; $\mathcal{N}(i)$ represents all one-hop neighbors of node $i$ and $c_{ij} = \sqrt{|\mathcal{N}(i)|}\sqrt{|\mathcal{N}(j)|}$. $\sigma$ denotes non-linear activation like ReLU. 

Following in-domain message-passing, we aim to enhance the discriminability of node embeddings deteriorated by network heterophily. While existing works focus on addressing this problem within the task graph~\cite{botgnn2,botgnn5,botgnn_gc1}, we instead explore common knowledge about bots and human users shared across multiple source graphs. Particularly, we learn a cross-source-domain topology to connect each source node to more advantageous neighbors. Although they originate from different source domains than the central node, they likely share similar labels and patterns with it and are located nearby in the embedding space. Thus, their information could contribute to recovering the discriminability of central node embeddings via message-passing.

To start with, for any source node $i$ from source domain $p$, we identify its Top-K similar nodes from all source domains excluding $p$ based on cosine similarity between node embeddings, and we establish cross-domain edges between $i$ and each of these K nodes. Thus, a Top-K sparse topology is created. However, edges may contain sub-optimal cross-domain connections between bots and human users, as the feature-camouflage characteristic of social bots may make their embeddings similar to human users. To score the quality of these edges, we resort to the consistency of local subgraph structures. As social relations are more difficult to camouflage, the local subgraph structure, which reveals account behaviors, could be similar within users in the same class (i.e., bot or human user). However, such subgraph-similarity is challenging for GNNs to capture. Therefore, for amy cross-domain edge $E_{ij}^{p,q}$ connecting node $i$, $j$ from source domains $p$ and $q$, we explicitly assign value $e_{ij}^{p,q}$ to it by measuring structural similarity between 3-hop ego-networks $G^{p}(i)$, $G^{q}(j)$ centred at $i$ and $j$, respectively: 
\begin{equation}
\label{graphlet}
   e_{ij}^{p,q}=k(G^{p}(i), G^{q}(j))\in (0,1) 
\end{equation}
where $k$ indicates graphlet kernel which maps local subgraphs to fixed-dim vectors and calculates such embedding similarity.  

Consequently, we conduct message-passing on these cross-source-domain edges to compute enhanced embeddings $Z_C$, which is initialized with $Z_I$. We adopt a 1-layer Graph Attention Network (GAT) encoder $f_a$ for aggregating beneficial information from cross-domain neighbors~\cite{gat}. Generally, the cross-source-domain message-passing could be represented as:

\begin{equation}
\label{crossdomainMP}
     Z^{D,(l)}_{C,i} = \sigma(b^c + \sum_{j\in\mathcal{N}(i)\cup i}\alpha_{ij}Z_{C,j}^{D',(l)}W^c) 
\end{equation}
where $D,D'\in \{S_1,\cdots, S_m\}$ and $D\neq D'$. $\mathcal{N}(i)$ refers to cross-domain neighbors of node $i$ from the other source domains. Attention coefficient $\alpha_{ij}$ (simple for $\alpha_{ij}^{D,D'}$) is computed by:
\begin{equation*}
\begin{aligned}
 \alpha_{ij} &= \frac{exp(s_{ij})}{\sum_{h\in \mathcal{N}(i)}exp(s_{ih})} \\
    s_{ij} &= \sigma'(a^c[z_{C,i}^{D,(l)}W^c\Vert z_{C,j}^{D',(l)}W^c\Vert e_{ij}^{D,D'}W^c_e])
\end{aligned} 
\end{equation*}
where $b^c$, $W^c$, $a^c$ and $W^c_e$ are parameters of GAT, $\sigma'$ represents LeakyReLU activation.

As the aggregated information from cross-domain neighbors might encompass diverse data distributions, and the updated node embeddings may be noisy. To alleviate this problem and further stabilize adaptation, we align the latent embedding spaces of source domains with a pairwise regularization $Reg^p$:
\begin{equation}
\label{sourcepair}
    Reg^p=\sum_i\sum_{j, j\neq i}Disc(Z^{S_i}_C,Z^{S_j}_C)
\end{equation}
where $Disc$ indicates discrepancy measurement (e.g., Maximum Mean Discrepancy). 

Finally, the overall embedding $z_i$ for any source node $i$ is obtained by aggregating embeddings from both in-domain ($Z_{I,i} \in \mathbb{R}^h$, Eq.~\ref{indomainMP}) and cross-domain ($Z_{C,i}\in \mathbb{R}^h$, Eq.~\ref{crossdomainMP}) channels, which enhances the discriminability of node embeddings while accurately encoding the inherent knowledge within each source domain:
\begin{equation}
\label{concat}
        z_i = \beta_1 Z_{I,i}+\beta_2 Z_{C,i} 
\end{equation}
where coefficients $\beta_1$ and $\beta_2$ are computed by applying a linear transformation matrix $W^{cat}$ mapping from $\mathbb{R}^{2h}$ to $\mathbb{R}^2$:
\begin{equation*}
    \beta_{1(2)} = \frac{exp(W^{cat}[Z_{I,i}\Vert Z_{C,i}])_{1(2)}}{\sum_{h\in\{1,2\}}exp(W^{cat}[Z_{I,i}\Vert Z_{C,i}])_h}
\end{equation*}

\subsection{Selective Multi-Source Transfer}
Following the network embedding via Eq.~\ref{concat}, we selectively transfer knowledge from multiple source graphs. The key idea is measuring the relevance between each source and target graph, i.e., $(G^{S_k},G^T)$. In this section, we first introduce a trivial multi-source adaptation pipeline to learn network-invariant and label-related embeddings with adversarial domain adaptation technique, then we discuss how to improve it in multi-source tasks. 

First of all, we use available labels in all source domains to train a shared classifier $g$, which will let source domain embeddings learn label-related knowledge:
\begin{equation}
\label{sup}
    L^{sup}=-\sum_{k}\frac{1}{n^{k}}\sum_{i=1}^{n^{k}}y_i^{S_k}log(g(f(z_i^{S_k})))
\end{equation}
where $i$ indicates the $i^{th}$ node, $S_k$ is the $k^{th}$ source domain, $n^{k}$ is the number of training samples from source graph $G^{S_k}$, $y_i^{S_k}$ is groundtruth label of node $i$ and $g(f(z_i^{S_k}))$ is the predicted logits.

Following the supervised training on source graphs, we eliminate domain discrepancy by modelling it with Wasserstein-1 distance, which has been proven to be effective in previous studies~\cite{wgan,specgda}. Consequently, we could learn network-invariant embeddings and transfer label-related knowledge from source to target graphs. Generally, the first Wasserstein distance between the source and target distributions of node embedding $Z^{S_k}$ and $Z^T$ is formulated as:
\begin{equation}
\label{eq_wdist}
    W_1(Z^{S_k}, Z^T) =  \sup_{||d||_{L} \leq 1} \mathbb{E}_{z \sim Z^{S_k}} [d(z)] - \mathbb{E}_{z' \sim Z^T} [d(z')]
\end{equation}
where $||d||_{L} \leq 1$ is Lipschitz continuity constraint, and $d$ is domain critic outputting a real number for any input node embedding. In practice, we approximate Eq.~\ref{eq_wdist} by maximizing the following domain critic loss:
\begin{equation}
\label{wdisc}
    L^{d} = \sum_k\frac{1}{n^{S_k}}\sum_{i}^{n^{S_k}}d(z^{S_k}_i)-\frac{1}{n^{T}}\sum_{j}^{n^{T}}d(z^{T}_j) 
\end{equation}

Therefore, by regarding the encoder $f$ from network embedding as generator and the domain critic $d$ as discriminator, we solve the following minimax problem:
\begin{equation*}
\min_{\Theta_f,\Theta_{oth}}\max_{\Theta_d} L^{d}
\end{equation*}
where $\Theta_f$, $\Theta_d$ denote encoder and classifier parameters, and $\Theta_{oth}$ includes all other model parameters including $f_a$ and $W^{cat}$.

In the above trivial settings, all sources are equally adapted to the target. Given the varying degrees of relevance between each source domain and the task, assigning equal weight to sources during supervised training and adaptation may either introduce noisy and irrelevant information from low-relevant sources or neglect beneficial knowledge from high-relevant sources, potentially impairing adaptation performance. To address this problem, we compute domain-level similarity $w^k$ between each source graph $G^{S_k}$ and target graph $G^T$. Subsequently, we employ a selective knowledge transfer strategy:
\begin{equation}
\label{loss}
    \begin{aligned}
        L^{sup}&=-\sum_{k}\frac{w^k}{n^{k}}\sum_{i=1}^{n^{k}}y_i^{S_k}log(g(f(z_i^{S_k}))) \\
        L^{d} &= \sum_k w^{k}(\frac{1}{n^{k}}\sum_{i}^{n^{k}}d(z^{S_k}_i)-\frac{1}{n^{T}}\sum_{j}^{n^{T}}d(z^{T}_j))
    \end{aligned}
\end{equation}

Generally, the weighted $L^{sup}$ and $L^{d}$ encourage the model to learn bot-related patterns and transfer more knowledge from source graphs that are more similar to target graph, respectively. 

The domain-level weight $w^k$ simultaneously models semantic and structural similarities between the source and target domains. First of all, as the output by domain discriminator $d$ directly measures similarity between latent embedding spaces, we take it into account and compute semantic similarity score $w^{k,d}\in (0,1)$:
\begin{equation*}
    w^{k,d} = \exp(-\vert \frac{1}{n^{k}}\sum_{i}^{n^{k}}d(z^{S_k}_i)-\frac{1}{n^{T}}\sum_{j}^{n^{T}}d(z^{T}_j)\vert)
\end{equation*}

Moreover, measuring the consistency of user behaviors from different domains is equally important because users usually exhibit different social behaviors on platforms with varying functionalities. These social behaviors are characterized by social relationships. Therefore, we apply graphlet kernel to compute structural similarity score $w^{k,g}\in (0,1)$ between topologies of each source graph $G^{S_k}$ and target graph $G^T$: $w^{k,g}=k(G^{S_k}, G^T)$, which is similar with Eq.~\ref{graphlet}. Consequently, $w^k$ is computed from both semantic and structure channels:
\begin{equation}
\label{eq::wk}
    w^k=\gamma w^{k,d} +(1-\gamma)w^{k,g} 
\end{equation}
where balancing coefficient $\gamma$ is set manually. 

\subsection{Anomaly Refinement}
In the inference stage, as source classifier $g$ may under-fit on the target graph due to noisy source labels, we refine the target predictions from $g$ with ``free" target domain knowledge. Specifically, we adopt the node-level heterophily score of each target node, which is defined from raw feature space:
\begin{equation*}
    score_i = 1-cosine(x_i,\frac{1}{\vert\mathcal{N}(i)\vert}\sum_{j\in\mathcal{N}(i)}x_j)
\end{equation*}
where $cosine$ indicates cosine similarity, $x_i$ and $x_j$ are raw node features. A higher value of $score_i$ indicates greater diversification of node $i$ from its neighbors, which suggests a potential bot. Therefore, during the inference stage, prediction logits for the target graph are refined by:
\begin{equation}
\label{refine}
    y_i^{pred}=\lambda g(f(z_i)) + (1-\lambda)score_i
\end{equation}
where $g(f(z_i))$ represents prediction logits output by GNN encoder $f$ and source classifier $g$. Hyper-parameter $\lambda$ is set manually.

\subsection{Training, Inference, and Complexity}
Putting all the ingredients together, i.e., Eq.~\ref{sourcepair} and~\ref{loss}, the overall loss of the proposed \textit{BotTrans} is as follows:
\begin{equation}
\label{final-loss}
    \min_{\Theta_f,\Theta_g,\Theta_{oth}}\{L^{sup}+\eta_1Reg^p+\max_{\Theta_d}\eta_2L^d\}
\end{equation}
where $\eta_1$ and $\eta_2$ are balancing coefficients. 

After training, we employ the encoder $f$ and classifier $g$ to infer the probabilities that target nodes are social bots. These logits are further refined via Eq.~\ref{refine}.  

The additional complexity of \textit{BotTrans} mainly stems from the CSD-MP and SMST module. In particular, SMST incurs a relatively low cost as it depends solely on discriminator outputs and pre-processed graph kernel scores (offline). The CSD-MP module introduces additional complexity through an extra GAT layer, $f_a$. Given $m$ subgraphs from sources, each containing $|v|$ nodes, the cross-source-domain topology then consists of $m|v|$ nodes and $Km|v|$ edges. Consequently, the complexity of $f_a$ is approximately $O(m|v| dd' + Km|v| d)$, where $d$ and $d'$ represent the input and output feature dimensions. Therefore, the subgraph sampling strategy and a small $K$ keep the whole scale acceptable.

\section{Experiments}

We evaluate \textit{BotTrans} to address the following research questions:
\begin{description}[leftmargin=0pt]
\item[RQ1] How does \textit{BotTrans} perform compared to baselines?
\item[RQ2] How does each module of \textit{BotTrans} contribute to the overall effectiveness?
\item[RQ3] Can \textit{BotTrans} stably excel when the number of source domains grows or source labels are limited?
\item[RQ4] How sensitive is \textit{BotTrans} to hyper-parameters?
\end{description}

\subsection{Experimental Settings}
\label{exp-setting}
\textbf{Real-World Dataset.} As the majority of public social bot datasets originate from Twitter, we mainly adopt three real-world, well-established datasets TwiBot-22, TwiBot-20 and MGTAB~\cite{Twibot20,feng2022twibot,MGTAB}. In particular, we build ten domains based on ten networks from these datasets. In addition to non-overlapping nodes and edges, users in different sub-networks have different social interests and their neighborhoods reflect distinct aspects of the whole Twitter network, leading to obvious feature- and structure- shifts across sub-networks. Thus, it is an effective strategy to test model transferability. \emph{To better focus on the two challenges proposed in this paper, we made a simplified assumption that the proportions of bots and human users are approximately balanced.}

\noindent  \textbf{Baselines.}  \emph{Inductive GNNs:} GCN~\cite{gcn} and GraphSAGE~\cite{sage} are trained on source domains and applied to the target domain without adaptation. \emph{Graph out-of-distribution (OOD) generalization:} We adopt EERM~\cite{graphood1} and ISGIB~\cite{graphood2}, which are first trained on source domains and subsequently used to make predictions directly on the target domain. \emph{Single-source Graph Domain Adaptation:} We adopt ADAGCN~\cite{adagcn}, GRADE~\cite{grade} and UDAGCN~\cite{udagcn}. \emph{Single-source Cross-Domain Graph Anomaly Detection:} We adopt ACT~\cite{act} and COMMANDER~\cite{cmd}. We use trivial multi-source settings for the above two categories, in which all sources are equally adapted. \emph{Multi-Source Domain Adaptation:} we adopt $M^3$SDA~\cite{msds_m3sda} and replace the encoder with GCN. \emph{Multi-source Graph Transfer Learning:} we adopt MSDS~\cite{msds} and NESTL~\cite{nestl}, both suitable for tasks involving multiple source graphs and an unlabelled target graph. \emph{Unsupervised Graph Anomaly Detection:} we adopt DGI+Local Outlier Factor (LOF), DGI+Isolation Forest (IF), CoLA~\cite{CoLA}, and CONAD~\cite{conad}. Specifically, DGI+LOF and DGI+IF indicate we firstly learn node embeddings with deep graph infomax~\cite{dgi} and apply unsupervised outlier detectors to discover social bots. 

\noindent  \textbf{Evaluation Protocol.} To build transfer tasks, we treat any domain as target domain $G^T$ (unlabeled) and select from the remaining domains as source domains (fully labeled). For efficiency, we first fix the number of source domains ($m$) at 2, 4, and 9, then build transfer tasks like: $\mathbf{G^{S_1},\cdots,G^{S_m}\rightarrow G^T}$. Subsequently, when $m$=2 or $m$=4, we fix two transfer tasks for each target domain: the transfer from the $m$ most relevant source domains (High-$m$) and the $m$ least relevant source domains (Low-$m$). Such relevance is assessed through preliminary experiments in Figure~\ref{figidea}, where a higher result indicate higher domain relevance. When $m$=9, all the remaining nine domains are used as sources. In total, each model is evaluated over twenty transfer tasks ($m$=2), twenty tasks ($m$=4), and ten tasks ($m$=9), respectively. This strategy efficiently assesses the model's knowledge transfer capability under different settings. 

\noindent  \textbf{Implement Details.} In \textit{BotTrans}, we use 2-layer GCN as in-domain encoder, 2-layer MLP as classifier and 1-layer MLP as domain discriminator, which are kept the same for baseline models. The graphlet kernels are implemented with Orca~\cite{orca}, which could be fast pre-processed before training. $\gamma$ and $\lambda$ are set a priori to 0.5, and balancing coefficients $\eta_1$, $\eta_2$ are both set to 1. We use top-5 neighbors to construct cross-source-domain topology. 
We train \textit{BotTrans} in a mini-batch manner with batch size equals 64 for all domains. For unsupervised methods, we use a prior contamination rate 0.5. We adopt the Adam optimizer with learning rate of 1e-4 and weight decay of 1e-5. Each experiment is repeated 5 times.

\subsection{Detection Effectiveness and Efficiency}
To answer \textbf{RQ1}, we conducted extensive experiments under various transfer settings and summarize the results in Table~\ref{tab:main} and~\ref{tab::unsup}. 

\noindent \textbf{Compared to supervised methods:} Generally, \textit{BotTrans} consistently outperforms and achieves significant performance improvements of around 3\% in both F1 and AUC. Remarkably, in the Low-$m$ transfer tasks with less similar source domains, \textit{BotTrans} maintains superior performance, unlike other baselines like AdaGCN or $M^3$SDA which exhibit a performance drop of up to 8\% from High-$m$ to Low-$m$ tasks. This demonstrates \textit{BotTrans}'s robust capability in transferring knowledge even under challenging conditions. Furthermore, key observations from Table~\ref{tab:main} include: 1) \textit{BotTrans} significantly surpasses single-source GDA models and multi-source graph transfer learning models, highlighting the effectiveness of our approach in alleviating source network heterophily and embedding over-smoothing, and selective knowledge-transfer strategy when leveraging multiple source domains, 2) Graph OOD models typically underperform in multi-source domain tasks. This limitation is likely attributed to the absence of specific knowledge of target domain, which hinders the model's ability to bridge the distribution gap between source and target domains, and 3) CDAD models underperform in social bot detection, likely due to their inability to adequately transfer bot-related knowledge and limited generalizability to the social bot detection task. 

\begin{table}
\centering
\caption{Multi-source transfer results (second best underlined). For 2 or 4 sources, we report the averaged F1-score over all 20 tasks and the High/Low-$m$ F1-score over 10 corresponding tasks. For 9 sources, we report the averaged F1-score over all 10 tasks. We also report AUC-ROC scores averaged over all tasks.}
\label{tab:main}
\resizebox{0.8\linewidth}{!}{%
\begin{tabular}{ccccc|cccc|cc} 
\toprule
          & \multicolumn{4}{c|}{2 sources}                                        & \multicolumn{4}{c|}{4 sources}                                        & \multicolumn{2}{c}{9 sources}      \\ 
\cline{2-11}
          & F1-avg          & \begin{tabular}[c]{@{}c@{}}High-m\\F1-avg\end{tabular}   & \begin{tabular}[c]{@{}c@{}}Low-m\\F1-avg\end{tabular}    & AUC-avg         & F1-avg          & \begin{tabular}[c]{@{}c@{}}High-m\\F1-avg\end{tabular}   & \begin{tabular}[c]{@{}c@{}}Low-m\\F1-avg\end{tabular}    & AUC-avg         & F1-avg          & AUC-avg          \\ 
\midrule
GCN       & 0.6463          & 0.6904          & 0.6021          & 0.7047          & 0.6738          & 0.6991          & 0.6485          & 0.7245          & 0.7054          & 0.7544           \\
GraphSAGE & 0.6486          & 0.7116          & 0.5856          & 0.7187          & 0.6989          & 0.7156          & 0.6853          & 0.7552          & 0.7098          & 0.7802           \\
EERM      & 0.6308          & 0.6686          & 0.5931          & 0.6842          & 0.6464          & 0.6525          & 0.6403          & 0.7164          &  0.6767          & 0.7482           \\
ISGIB     & 0.6521          & 0.6880          & 0.6162          & 0.7265          & 0.6786          & 0.6933          & 0.6640          & 0.7523          &  0.688           & 0.7628           \\
AdaGCN    & 0.7206          & \uline{0.7682}  & 0.6821          & 0.7501          & \uline{0.7536}  & \uline{0.7644}          & 0.7246          & 0.7748  & \uline{0.7768}  & \uline{0.8246}   \\
Grade     & \uline{0.7280}  & 0.7258          & \uline{0.7202}  & 0.7417          & 0.7338  & 0.7460  & 0.7224          & 0.7382          & 0.6544          & 0.7563           \\
UDAGCN    & 0.7090          & 0.7344          & 0.6836          & \uline{0.7806}  & 0.7481  & 0.7579  & \uline{0.7483}  & \uline{0.7892}  & 0.762           & 0.824            \\
Commander & 0.6501          & 0.6611          & 0.6411          & 0.7060          & 0.6546          & 0.6894          & 0.6411          & 0.7370          & 0.6530          & 0.7543           \\
ACT       & 0.4148          & 0.4100          & 0.4197          & 0.4794          & 0.4270          & 0.4218          & 0.4313          & 0.4908          & 0.3960          & 0.4761           \\
$M^{3}$SDA     & 0.7014          & 0.7415          & 0.6613          & 0.7519  & 0.7098          & 0.7227          & 0.6993          & 0.7409          & 0.7014          & 0.7309           \\
MSDS      & 0.6579          & 0.6606          & 0.6552          & 0.6713          & 0.6572          & 0.6650          & 0.6494          & 0.6793          & 0.6732          & 0.7043           \\
NESTL     & 0.5427          & 0.6058          & 0.4796          & 0.7261          & 0.5502          & 0.6188          & 0.4941          & 0.7261          & 0.5283          & 0.7262           \\ 
\hline
BotTrans  & \textbf{0.7569} & \textbf{0.7819} & \textbf{0.7319} & \textbf{0.7983} & \textbf{0.7851} & \textbf{0.7926} & \textbf{0.7789} & \textbf{0.8260} & \textbf{0.8000} & \textbf{0.8524}  \\
\bottomrule
\end{tabular}
}
\end{table}

\noindent  \textbf{Compared to unsupervised methods:} we compare model performance on each domain $d$ with \textit{BotTrans}'s result, which is given by the best result among all multi-source adaptation tasks with $d$ as target domain. The results in Table~\ref{tab::unsup} demonstrate that even the best baseline DGI+IF is significantly inferior to \textit{BotTrans}, with a performance decrease of 18.34\% $\sim$ 36.82\%. These results prove the necessity of knowledge transfer for detecting social bots.

\begin{table}
\centering
\caption{Unsupervised experiment results (F1-score). `$\uparrow$` indicates the performance gain of \textit{BotTrans} against the best unsupervised model.}
\label{tab::unsup}
\resizebox{0.8\linewidth}{!}{%
\begin{tabular}{c|cccccccccc} 
\hline
                               & 0               & 1               & 2               & 3               & 4               & 5               & 6               & 7               & 8               & 9                \\ 
\hline
DGI+LOF                        & \uline{0.5471}          & \uline{0.5455}          & \uline{0.5392}          & \textbf{0.5582} & \uline{0.6017}          & 0.4808          & \uline{0.5106}          & \uline{0.4928}          & \uline{0.5190}          & \uline{0.5157}           \\
DGI+IF                         & \textbf{0.5797} & \textbf{0.6477} & \textbf{0.6572} & \uline{0.5484}          & \textbf{0.6966} & \textbf{0.6215} & \textbf{0.5280} & \textbf{0.5254} & \textbf{0.5510} & \textbf{0.5170}  \\
CoLA                           & 0.4824          & 0.4975          & 0.5022          & 0.5118          & 0.5012          & \uline{0.4971}          & 0.4835          & 0.4920          & 0.4894          & 0.5005           \\
CONAD                          & 0.4207          & 0.4419          & 0.4498          & 0.4506          & 0.4383          & 0.4339          & 0.4802          & 0.4128          & 0.4134          & 0.3945           \\ 
\hline
\textit{BotTrans} & 36.82\%$\uparrow$         & 24.60\% $\uparrow$        & 20.66\%$\uparrow$         & 33.30\% $\uparrow$        & 18.48\%$\uparrow$         & 20.24\%$\uparrow$         & 18.34\%$\uparrow$         & 31.93\%$\uparrow$         & 27.76\%$\uparrow$         & 32.20\%$\uparrow$          \\
\hline
\end{tabular}
}
\end{table}

\begin{figure}[ht]
  \centering
  \includegraphics[width=0.7\textwidth]{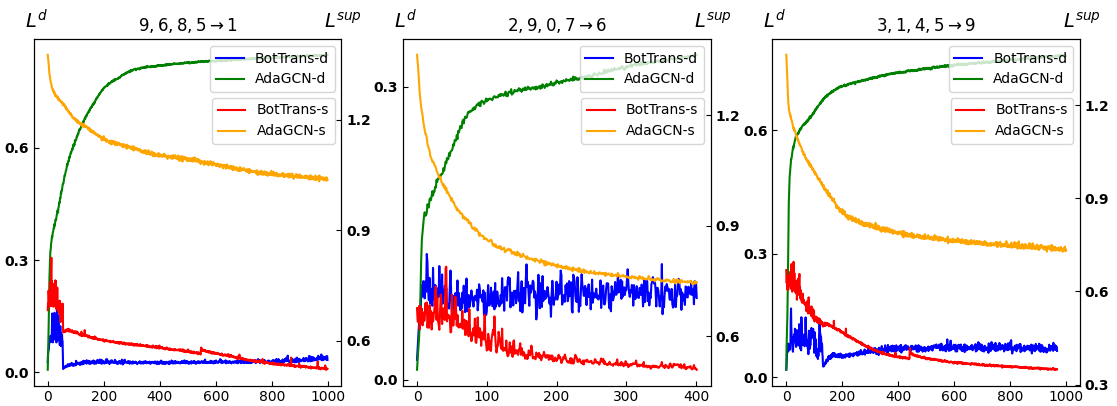}
  \caption{Training with $L^d$ and $L^{sup}$. X-axis: training epoch. Y-axis: $L^d$ (left, maximized) and $L^{sup}$ (right, minimized). d and s are short for $L^d$ and $L^{sup}$.}
  \label{figloss}
\end{figure}

\noindent \textbf{Effectiveness analysis:} Guided by the theory of generalization bounds, we explore the reasons behind our model's effectiveness by experimentally assessing its capability to minimize the supervised loss within the sources and to reduce discrepancies between source and target domains. Thus, we plot the training process in three representative transfer tasks for \textit{BotTrans} with another strong baseline AdaGCN in Figure~\ref{figloss}. Key findings include: 1) \textit{BotTrans} achieves lower source training error $L^{sup}$, indicating the CSD-MP module mitigates the network heterophily and over-smoothing problem in source domains, and 2) \textit{BotTrans} achieves lower domain divergence $L^d$, indicating the SMST module proposes better adaptation strategy in multi-source tasks. 

\noindent \textbf{Efficiency analysis:} We compare with strong baselines in Table~\ref{tab:main}. \textit{Time Complexity}: \textit{BotTrans}'s running time matches AdaGCN but is 2$\times$M$^3$SDA's, 3$\times$Commander's, and 4$\times$Grade's. However, it offers up to 19\% performance boost. \textit{Spatial Complexity}: \textit{BotTrans} only requires an additional few hundred megabytes in memory and GPU usage, which is mainly introduced by CSD-MP. Overall, \textit{BotTrans} achieves a balance between effectiveness and efficiency.

\subsection{Ablation Study}
To address RQ2, we perform ablation studies by incrementally adding key model components, with results summarized in Table~\ref{tab:ablation}. We begin with a variant, Base, which uses a simple GCN encoder (Eq.\ref{indomainMP}) and single-source adaptation components (Eq.\ref{sup} and \ref{wdisc}). Next, we implement CSD-MP (w/o. edge feature), based on Base, but with cross-source-domain topology excluding edge features. We then sequentially introduce three additional variants. The results demonstrate that progressively adding components improves performance. Specifically, the full CSD-MP module yields a performance increase of 1.6\%$\sim$2.9\% across various transfer tasks, which we attribute to its effective mitigation of the network heterophily issue. Additionally, SMST enhances performance by 1.1\%$\sim$2.7\%, highlighting the importance of selective knowledge adaptation. Finally, the AR module contributes an improvement of 0.8\%$\sim$1.4\%, addressing potential issues with noisy source labels.

\begin{table}
\centering
\caption{Ablation study results (F1). When there are 2 or 4 sources, we report the results averaged over 20 tasks. When there are 9 sources we report the results averaged over 10 tasks.}
\label{tab:ablation}
\resizebox{0.6\linewidth}{!}{%
\begin{tabular}{l|c|c|c} 
\toprule
                            & 2 sources & 4 sources & 9 sources  \\ 
\midrule
Base                        & 0.7179    & 0.7239    & 0.7418     \\
+CSD-MP (w/o. edge feature) & 0.7253    & 0.7341    & 0.7466     \\
+CSD-MP                     & 0.7346    & 0.7521    & 0.7573     \\
+CSD-MP+SMST                & 0.7456    & 0.7747    & 0.7846     \\
+CSD-MP+SMST+AR             & 0.7536    & 0.7846       & 0.7980        \\
\bottomrule
\end{tabular}
}

\end{table}

\subsection{Stability}

\begin{figure}[ht]
  \centering
  \includegraphics[width=0.6\linewidth]{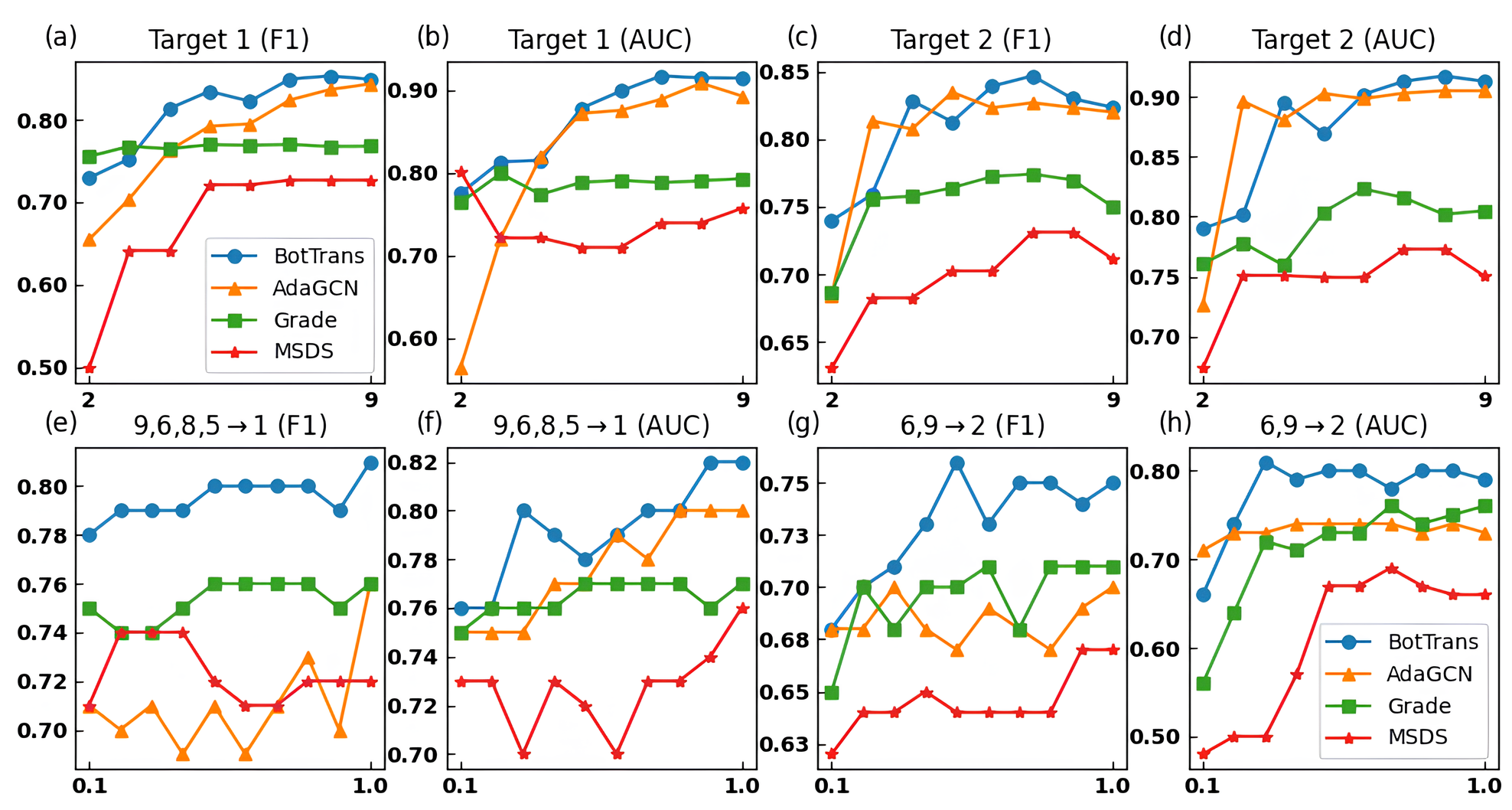}
  \caption{Increasing the number of source domains (a-d) and increasing the percentage of labeled source nodes (e-h). }
  \label{figstability}
\end{figure}

To address \textbf{RQ3}, we evaluate model stability under varying numbers of source domains and label utilization rates. This is important since, in practice, it can be difficult to obtain a large number of source domains with sufficient bot-related labels. A robust model should therefore perform consistently well regardless of these factors. We select strong baselines along with \textit{BotTrans}, fix two target domains(e.g., $G^1$ and $G^2$), and measure performance as the number of source domains increases from 2 to 9. As shown in Figure~\ref{figstability}, \textit{BotTrans} consistently outperforms baselines and benefits from more source domains. Next, we fix two representative tasks and vary the labeled source node ratio from 10\% to 100\%. As illustrated in the figure, \textit{BotTrans} achieves strong results across all labeling rates, demonstrating robustness and adaptability even with limited supervision.

\begin{figure}[ht]
  \centering
  \includegraphics[width=0.6\linewidth]{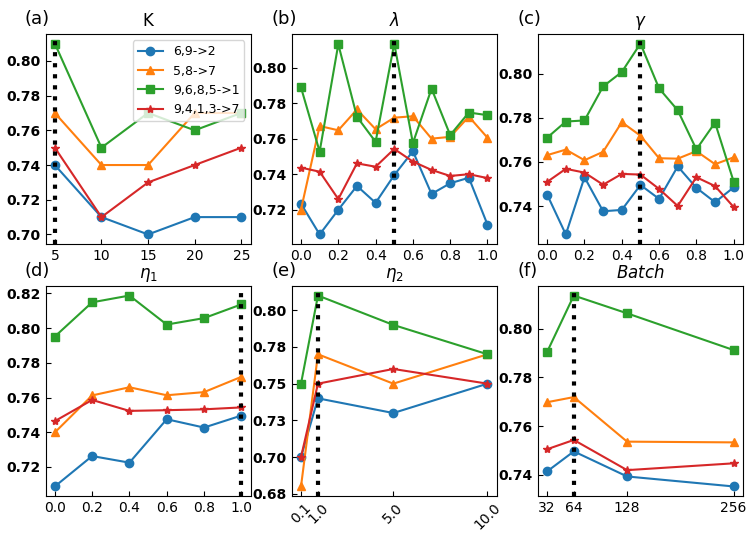}
  \caption{Hyper-parameters tuning, where default settings are marked with vertical dotted lines.}
  \label{figsensitivity}
\end{figure}

\subsection{Sensitivity}
We perform sensitivity analysis of \textit{BotTrans} on important hyper-parameters to answer \textbf{RQ4}, including cross-source-domain neighboring number $K$, balancing coefficients $\lambda$ from Eq.~\ref{refine} and $\eta_1$ from Eq.~\ref{final-loss}, and batch size. As Figure~\ref{figsensitivity} reveals, the default settings are adequate for most adaptation tasks, eliminating the need for extensive tuning. Besides, when $K$ is added up to 5 that enables stable model performance, further increasing $K$ tends to incorporate more noisy information from other source domains and deteriorate performance. Moreover, given that model performs better when values of $\gamma$ and $\lambda$ lie between 0 and 1 compared to when they are set to 0 or 1, we can also validate the effectiveness of domain-level similarity score computing (Eq.~\ref{eq::wk}) and consistency-score refinement (Eq.~\ref{refine}).

\section{Conclusion}
To alleviate the label scarcity issue in detecting social bots with GNN-based methods, we introduce \textit{BotTrans}, a novel MSGDA model designed to transfer extensive bot-related knowledge from multiple social networks. It comprises three key modules focused on mitigating network heterophily in source domains, selectively transferring relevant knowledge from source domains relevant to the target domain, and refining target detection results. Experimental studies prove that \textit{BotTrans} outperforms all SOTA baselines in unlabeled detection tasks. In the future, we will explore more approaches, including pre-trained graph models and large language models. 

\begin{credits}
\subsubsection{\ackname} This work was supported by the National Key Research and Development Program of China under Grant No. 2022YFB3103704, the National Natural Science Foundation of China under Grant Nos. 62372434, U21B2046, and 62302485, CAS Special Research Assistant Program and the Key Research Project of Chinese Academy of Sciences (No. RCJJ-145-24-21).
\end{credits}
%


\end{document}